# Unsupervised Feature Learning with K-means and An Ensemble of Deep Convolutional Neural Networks for Medical Image Classification

Euijoon Ahn[1], Ashnil Kumar[1], Dagan Feng[1, 4], Michael Fulham[1, 2, 3], and Jinman Kim[1]

[1] School of Computer Science, University of Sydney, Australia
[2] Department of Molecular Imaging, Royal Prince Alfred Hospital, Australia
[3] Sydney Medical School, University of Sydney, Australia
[4] Med-X Research Institute, Shanghai Jiao Tong University, China
`jinman.kim@sydney.edu.au`

**Abstract.** Medical image analysis using supervised deep learning methods remains problematic because of the reliance of deep learning methods on large amounts of labelled training data. Although medical imaging data repositories continue to expand there has not been a commensurate increase in the amount of annotated data. Hence, we propose a new unsupervised feature learning method that learns feature representations to then differentiate dissimilar medical images using an ensemble of different convolutional neural networks (CNNs) and *K*-means clustering. It jointly learns feature representations and clustering assignments in an end-to-end fashion. We tested our approach on a public medical dataset and show its accuracy was better than state-of-the-art unsupervised feature learning methods and comparable to state-of-the-art supervised CNNs. Our findings suggest that our method could be used to tackle the issue of the large volume of unlabelled data in medical imaging repositories.

**Keywords:** Unsupervised Feature Learning, Clustering, Convolutional Neural Network, Modality Classification.

## 1 Introduction

Supervised deep learning methods allow the derivation of image features for a variety of image analysis problems using underlying algorithms and large-scale labelled data [1]. In the medical imaging domain, however, there is a paucity of labelled data due to the cost and time entailed in manual delineation by imaging experts, inter- and intra-observer variability amongst these experts and then the complexity of the images themselves where there may be many different appearances based on for instance, bone and soft tissues windows on computed tomography (CT) and different sequences on Magnetic Resonance (MR) imaging.

Researchers have employed many different approaches to help solve these challenges including deep learning with transferable knowledge across different domains using natural images as a general feature extractor [2, 3] and fine-tuning generic knowledge with a relatively smaller amount of labelled image data. Other approaches use unsupervised feature learning [2] where the aim is to learn image representations that can better reconstruct training samples. The process of learning representative image features from unlabelled data, however, is non-trivial and unsupervised feature learning is inferior to supervised methods. Alternatively, image clustering has been used to separate dissimilar images. Clustering is central to unsupervised feature learning methods and has been applied to image segmentation [3-5] and classification [6].

Our aim was to develop an unsupervised approach to learn features from medical images, exploiting clustering to generate surrogate labels that could be used to maximise the separation between dissimilar images. We investigated the accuracy of our method on a public medical dataset and compared it to other state-of-the-art unsupervised and supervised methods. We suggest that our method could potentially provide an opportunity to learn image features in the large-volume of unlabelled data in medical imaging repositories.

### 1.1 Related Work

In unsupervised feature learning methods, image features are generally learned using algorithms such as sparse coding (SC) [7], the stacked sparse auto-encoder (SSAE) [8] and Restricted Boltzmann Machines (RBMs). These algorithms, however, have been limited to learning low-level image features. Recently, kernel learning coupled with deep learning architectures [9, 10] provided promising results in image classification [11] and retrieval [2, 12]. These methods learned image features in a feed-forward manner, i.e., weights learned from earlier layers are fixed, and so do not take advantage of iterative learning.

Many clustering algorithms are based on *K*-means, Gaussian mixture model, spectral clustering and agglomerative clustering. A notion of similarity, based on data representation in a feature space and distance metrics, is critical to clustering algorithms. Clustering algorithms are easy to use however, they are problematic when the quality of data representation in a feature space is poor. Recently, several investigators addressed this by using deep learning architectures to extract more representative image features [13, 14]. Xie et al. [13] used a stacked auto-encoder to initialise clustering assignments and leaned image features via spectral clustering. Similarly, Yang et al. [15] iteratively learned image features using a CNN and clustered them using agglomerative clustering. These approaches relied on one specific deep learning architecture.

### 1.2 Contribution

We propose an unsupervised feature learning method that learns image features and cluster assignment (i.e., clustering) simultaneously in an end-to-end fashion. Our method uses the *K*-means clustering algorithm and leverages the subsequent distinct clusters as surrogate labels to learn the parameters (i.e., weights) of deep CNNs. Our conceptual idea is shown in Fig 1, where we propose an ensemble of different unsupervised CNN architectures to improve feature representations in medical images.

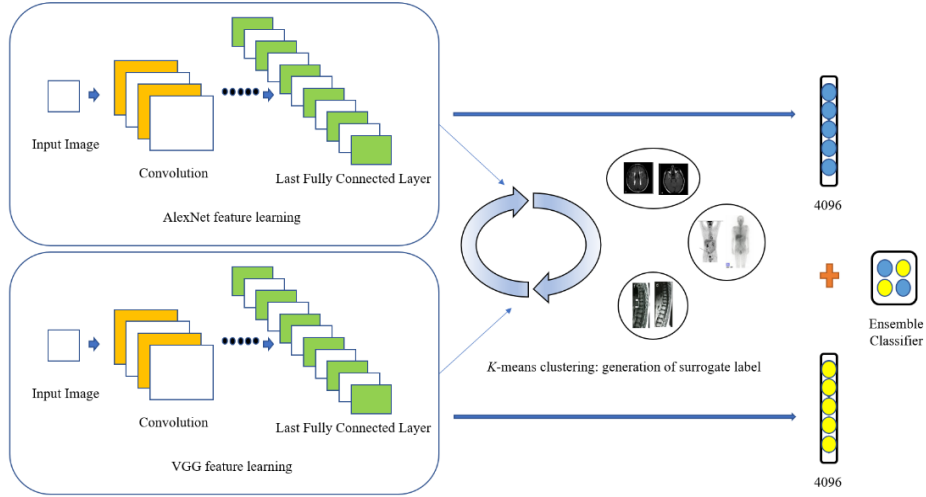

**Fig. 1.** Overview of our unsupervised feature learning method.

## 2 Methods

### 2.1 Unsupervised Feature Learning via *K*-means and CNNs

*K*-means iteratively identifies nearby features based on the distances calculated from centroids and assigns them to the closest cluster. *K*-means is used to construct a dictionary $D \in \mathbb{R}^{d \times k}$ of $k$ vectors (i.e., centroids) such that a data vector $x_i \in \mathcal{R}^d, i = 1, \ldots, m$ can be mapped to a code vector $s_i$ that minimises the error in reconstruction, which is defined as follows:

$$\min_{D\ s} \sum_{i=1}^{N} \|Ds_i - f(x_i, w)\|_2^2 \quad (1)$$

$$\text{subject to } \|s_i\|_0 \leq 1, \forall_i \quad (2)$$

$$\text{and } \|D_j\|_2 = 1, \forall_i \quad (3)$$

where $x_i$ is the image input, $f(x_i, w)$ is the CNN function that computes image features, given input $x_i$ and weights $w$, and $D_j$ is the $j$th column of the dictionary $D$. The aim here is to learn a dictionary $D \in \mathbb{R}^{d \times k}$ and a new code vector of $s_i$, which then allows to the reconstruction of the original CNN features. Solving Equation (1) provides a set of optimal cluster assignments, $\tilde{y}_i$ which we used as surrogate labels for learning parameters of CNNs in our ensemble. The parameters of each CNN were then learned by optimising the following problem:

$$\min_{w} \frac{1}{N} \sum_{i=1}^{N} \mathcal{L}(f(x_i, w), \tilde{y}_i) \qquad (4)$$

We set $\mathcal{L}$ to the cross entropy loss, and is minimised using stochastic gradient descent (SGD) as in a standard CNN backpropagation algorithm. The training is an end-to-end iterative process: $K$-means clustering using CNN features produces surrogate labels (see Equation 1) and the parameters of the CNNs are then updated using the surrogate labels (see Equation 4). This process is repeated until the clustering and loss $\mathcal{L}$ become stable.

### 2.2 Data Pre-processing and Initialisation for Clustering

Data pre-processing - normalisation and whitening - is crucial to clustering [16]. Raw data tends to generate many highly correlated cluster centroids rather than the centroids that capture diverse characteristics across whole data [17]. So we used principal component analysis (PCA) whitening (256 dimensions) to remove these correlations, followed by $l_2$ normalisation to increase the local contrast.

The effective initialisation of the centroids in $K$-means is important to avoid empty clusters. Instead of initialising the centroids using randomly chosen examples from training data, we randomly initialised the centroids from a Normal distribution and then normalised them to unit length equal to 1. If a cluster still becomes empty, we reinitialised its centroids with random examples.

### 2.3 Deep Learning with CNNs

We used AlexNet [18] and VGG-16 [19] for the CNN architectures. The output of the last fully connected layer is a vector of 4096 dimensions. We used this as a feature vector for the subsequent clustering.

### 2.4 Linear SVM and An Ensemble of CNNs

We extracted the image features from the last fully connected layer from the learned CNNs (4096 dimensions) for linear SVM classifiers. We used the one-against-all linear SVM with a differentiable quadratic hinge loss [20], so that the training could be done with simple gradient-based optimisation methods. We used LBFGS with a learning rate of 0.1 and a regularization parameter of 1, consistent with the parameters specified by Yang et al. [20]. The outputs (prediction scores) from each CNN feature are then fused (averaging) to estimate the final classification results.

## 3  Experimental Setup

We used the medical Subfigure Classification dataset (SCd) from the Image Conference and Labs of the Evaluation Forum (ImageCLEF) 2016 competition [21]. The dataset has 6776 training and 4166 test images from 30 different imaging modalities with ground-truth annotations for both datasets. The SCd has various anatomical images from a variety of diseases and this diversity poses challenges for medical image analysis.

As the baseline, we compared our method to SC and Independent Component Analysis (ICA), which are established unsupervised feature learning methods. We also compared it to state-of-the-art unsupervised methods - SSAE, convolutional kernel network (CKN) [11] and a sparsity-based convolutional kernel network (S-CKN) [2] – and to supervised methods based on transfer-learned and fine-tuned CNNs [22]. We used the correctness of the predicted label – the Top 1 image accuracy – as the performance measure, which was adopted in recent CNN studies for the classification of medical images [22]. For all the unsupervised learned features (SC, ICA, SSAE, CKN and S-CKN), we used the same one-against-all linear SVM [20].

### 3.1  Implementation Details

The main parameter of *K*-means is the number of clusters (i.e., number of surrogate labels). We used an empirical process to discover appropriate number of clusters (10, 30, 50, 100, 150 and 200). We set $k = 50$ in our all experiments.

Our CNN was trained for 500 epochs with a batch size of 256 for AlexNet and 32 for VGG. The clustering assignments were updated every epoch and the initial learning rate was set to 0.05. We used random horizontal flips and crops for data augmentation and dropout, constant step size, momentum of 0.9 and $l_2$ penalisation of the weights on a GeForce GTX 1080 Ti GPU (11GB memory).

## 4  Results and Discussion

Our results are outlined in Tables 1 and 2. They show that our method had greater accuracy than other state-of-the-art unsupervised methods with a Top 1 accuracy of 74.51%. It also had a comparable accuracy (78.61%) with the supervised transfer-learned CNNs. The best performing methods was the ensemble of fine-tuned GoogLeNet and ResNet with an accuracy of 83.14%.

Our findings indicate that our unsupervised method was able to improve the feature representation of medical images by maximising the separation between dissimilar medical images using clustering coupled with CNNs. Our ensemble scheme further improved the feature representation (see Table 1), and we attribute this to its ability to learn and capture different types of representative image features.

The qualities of the features learned with conventional unsupervised feature learning methods such as SC and ICA were not as robust as that of SSAE. Different to SC

46

and ICA, SSAE learned image features in a hierarchical manner and hence had a higher accuracy (65.17%). Among all unsupervised feature learning methods, kernel-based CNNs such as CKN (68.22%) and S-CKN (70.99%) were the closest to our method. They were able to learn more discriminative features by learning image features in a unified reproducing kernel Hilbert space (RKHS). The kernel-based CNNs, however, only learn image features in a feed-forward manner and do not take advantage of the iterative scheme that we used in our method, i.e., simultaneous image representation learning and clustering assignments.

The top performing methods reported in the competition were all based on well-established supervised CNNs including AlexNet [22], VGG [23], GoogLeNet [22] and ResNet [24]. These CNNs were trained from scratch or fine-tuned with medical images. Our unsupervised method (accuracy of 74.51%) performed better than supervised VGG-like CNNs (65.31%) with over 9% improvement in modality classification.

Although our method improved the feature representations of medical images in an unsupervised fashion, the parameter $k$, i.e., number of clusters needs to be empirically derived. Generally, it is better to set a higher $k$ than the total number of actual class labels (see Table 2); setting it to a lower $k$ will force dissimilar images to stay in same cluster. It is possible that deeper CNNs such as GoogLeNet and ResNet may potentially provide different feature representations to learn more discriminative medical image features. We will explore such approaches as future work.

**Table 1.** The Top 1 Image accuracy results on the ImageCLEF16 dataset compared to state-of-the-art unsupervised and supervised methods.

| Type | Methods | Top 1 Accuracy (%) |
| --- | --- | --- |
| Unsupervised | Sparse Coding | 57.08 |
| Unsupervised | ICA | 58.79 |
| Unsupervised | Stacked Sparse Auto-encoder (2 layers) | 65.17 |
| Supervised | VGG-like CNN [23] | 65.31 |
| Unsupervised | Convolutional Kernel Network [11] | 68.22 |
| Unsupervised | Sparsity-based Convolutional Kernel Network [2] | 70.99 |
| Unsupervised | Our method with AlexNet architecture | 72.37 |
| Unsupervised | Our method with VGG architecture | 72.64 |
| Unsupervised | Our method with an ensemble of AlexNet and VGG | 74.51 |
| Supervised | Transfer-learned GoogLeNet + SVM [22] | 78.61 |
| Supervised | An ensemble of fine-tuned AlexNet and GoogLeNet [22] | 82.48 |
| Supervised | An ensemble of fine-tuned GoogLeNet and ResNet [24] | 83.14 |

**Table 2.** Results of Top 1 accuracy using VGG-16 with different numbers of clusters $k$

| $k$ | Top 1 Accuracy (%) |
|-----|--------------------|
| 10  | 70.64              |
| 30  | 72.42              |
| 50  | 72.64              |
| 100 | 72.35              |
| 150 | 71.82              |
| 200 | 71.50              |

## 5 Conclusion

Our unsupervised feature learning method jointly learned image feature representations and clustering using an ensemble of different CNNs in an end-to-end fashion. It outperformed other state-of-the-art unsupervised methods, and its accuracy was comparable to the supervised CNNs. We suggest that our method may provide an opportunity to learn large-volume of unlabelled data in medical imaging repositories.